\documentclass[journal]{IEEEtran}
\usepackage{caption}
\usepackage{tikz}
\usepackage{amssymb}
\usepackage{amsthm}
\theoremstyle{plain}

\usepackage{amsmath}
\usepackage{algpseudocode}
\usepackage{tabulary}
     
\usepackage{enumitem}

\usepackage{graphicx}
\usepackage{subcaption}
\usepackage[export]{adjustbox}

\usepackage{booktabs}
\usepackage{threeparttable}

\usepackage{adjustbox}
\usetikzlibrary{positioning, shapes.geometric}
\usepackage{xcolor}

\usepackage{mathtools}

\usepackage{makecell}
\newcolumntype{P}[1]{>{\centering\arraybackslash}p{#1}}

\usepackage{cite}
\usepackage{floatrow}
\usepackage{lipsum}
\usepackage{multicol}
\newcommand{\nonl}{\renewcommand{\nl}{\let\nl\oldnl}}

\usetikzlibrary{arrows,shapes,positioning,shadows,trees}

\usepackage{algorithm}

\usepackage{booktabs}
\usepackage{multirow}   
\usepackage{array}     
\usepackage{tabularx}   
\usepackage{threeparttable}

\tikzset{
  basic/.style  = {draw, text width=2cm, drop shadow, font=\scriptsize, rectangle},
  root/.style   = {basic, rounded corners=2pt, thin, align=center,
                   fill=white!30},
  level 2/.style = {basic, rounded corners=4pt, thin,align=center, fill=white!60,
                   text width=5em},
  level 3/.style = {basic, thin, align=left, fill=white!60, text width=5em}
}

\newcommand{\CASE}[1]{\STATE \textbf{case} #1\textbf{:} \begin{ALC@g}}
\newcommand{\ENDCASE}{\end{ALC@g}}

\newcommand{\DEFAULT}{\STATE \textbf{default:} \begin{ALC@g}}
\newcommand{\ENDDEFAULT}{\end{ALC@g}}
\newcommand{\DEFAULTLINE}[1]{\STATE \textbf{default:} }

\makeatletter
\g@addto@macro\normalsize{%
  \setlength\abovedisplayskip{6pt plus 2pt minus 2pt}%
  \setlength\belowdisplayskip{6pt plus 2pt minus 2pt}%
  \setlength\abovedisplayshortskip{4pt plus 2pt minus 2pt}%
  \setlength\belowdisplayshortskip{4pt plus 2pt minus 2pt}%
}
\makeatother
\usepackage[font=small,skip=4pt]{caption}

\ifCLASSINFOpdf
\else
\fi
\begin{document}
%
\title{Edge-AI-Driven Learning-to-Rank for Decentralized Task Allocation in Circular Smart Manufacturing}
%
%

\author{Mohammadhossein Ghahramani,~\IEEEmembership{Senior Member,~IEEE,} Yan Qiao,~\IEEEmembership{Senior Member,~IEEE,} \\ and Mengchu Zhou,~\IEEEmembership{Fellow,~IEEE}

\thanks{}
\thanks{}
\thanks{M. Ghahramani is with Birmingham City University, UK, (e-mail: mohammadhossein.ghahramani@bcu.ac.uk)}

\thanks{Y. Qiao is with the College of Computer and Information Science, Southwest University, Chongqing 400000, China, and also with IKAS Holdings (Beijing) Co., Ltd., Beijing 100000, China (e-mail: qiaoyan2026@swu.edu.cn).}

\thanks{M. C. Zhou is with the Helen and John C. Hartmann Department of Electrical and Computer Engineering, New Jersey Institute of Technology, Newark, NJ 07102, USA (e-mail: zhou@njit.edu).}

}

%
%

\markboth{}%
{Shell \MakeLowercase{\textit{et al.}}: Bare Demo of IEEEtran.cls for IEEE Journals}
%



\maketitle

\begin{abstract}
Task allocation in smart manufacturing systems must operate under decentralized decision-making, dynamic workloads, and shared-resource constraints. In circular manufacturing settings, these challenges are further intensified because tasks compete for reusable and capacity-constrained assets. Moreover, machine selection might affect processing energy. Although learning-based approaches have been explored for task allocation, improvements in predictive modeling do not necessarily translate into better allocation outcomes under decentralized negotiation. This work proposes an Edge-AI-driven decentralized task-allocation framework. We develop lightweight decision intelligence deployed at the machine level. It is developed progressively: first, a resource-aware heuristic establishes the decentralized bidding structure; a regression-based Edge-AI formulation then examines learned local bid approximation, and a compact autoencoder-regularized pairwise ranking model finally provides a learned correction to the analytical bid ordering. Each machine evaluates incoming tasks by using its processing capability, queue state, energy characteristics, and a compact signal representing contention over the reusable shared production asset. The framework is assessed via discrete-event simulation in scenarios characterized by high load and dependence on shared resources. Compared to the heuristic, the proposed ranking method increases completed tasks, reduces average tardiness, and lowers the deadline-miss rate, with statistically significant paired differences. Mean energy per completed task is also reduced. The results indicate that effective learning-assisted allocation depends not only on approximating local decision quantities, but also on shaping the relative preferences that determine negotiation outcomes. This ranking-aware decision support enables more effective operational coordination under high dependence on reusable shared assets, thereby supporting the resource-efficiency objectives of circular manufacturing.

\makeatletter{\renewcommand*{\@makefnmark}{}
\footnotetext{}\makeatother}

\end{abstract}


%
\IEEEpeerreviewmaketitle

\section{Introduction}\label{section.intro}
%
%
%
%

\IEEEPARstart{S}{mart} manufacturing systems are characterized by dynamic workloads, heterogeneous processing capabilities, and stringent operational constraints \cite{Ren2026Digital,Chen2025Workload,Ghahramani2021,Ghahramani2026CLAIRE, Ghahramani-Spatial}. In emerging circular manufacturing environments, conventional production is increasingly integrated with inspection, repair, reconditioning, and remanufacturing operations. These operations rely on the repeated use of shared production assets, such as tools, fixtures, inspection equipment, and rework facilities \cite{Narula2025Putting,Reyes2025Understanding}. In such settings, efficient task allocation is fundamental not only to system responsiveness, but also to the effective utilization of reusable production resources and processing energy. Classical approaches to task allocation are typically formulated as centralized optimization or scheduling problems. In these methods a global controller assigns tasks based on complete system information \cite{Huang2024Evolutionary,Li2023Multi,Li2025Decoupling,Lai2026Optimization,Deng2025Evolutionary}. Although such approaches can achieve high-quality solutions (under static or moderately dynamic conditions), their applicability is limited. This is due to scalability challenges, communication overhead, and the need for timely decision-making. These limitations have led to the development of decentralized task allocation mechanisms. In these approaches, machines act as autonomous agents and make local decisions based on partial information \cite{Peng2022DoSRA,Govoni2025Decentralized,Chen2026Surrogate}. Negotiation-based methods (particularly those utilizing bidding) represent a flexible and scalable paradigm \cite{Tang2024Efficient,Chen2025Optimal}. Within such systems, machines assess the incoming tasks according to local state variables and submit bids to inform the assignment process. This mechanism supports adaptive behavior without the need for global coordination.

Despite the advantages of decentralized negotiation-based approaches, they face fundamental challenges in capturing system-level interactions. In practical environments, machines often share limited resources. From a circular manufacturing perspective, these shared resources can be interpreted as reusable production components whose efficient utilization is critical. Purely local bidding strategies cannot fully account for these interactions. Consequently, they may lead to suboptimal allocations under high-load conditions or tight scheduling constraints. Recent advances have explored the integration of learning-based techniques into decentralized decision-making \cite{Chen2023Deep}. However, many existing approaches focus on predicting absolute performance measures, such as completion time or cost, which do not necessarily translate into improved allocation outcomes. This is because task assignment decisions are inherently comparative: they depend on the relative ordering of candidate machines instead of the absolute values of their evaluations. Consequently, learning formulations that are not aligned with this decision structure may fail to influence system behavior in a meaningful way.

To address these challenges, we propose an Edge-AI-driven decentralized task allocation framework based on ranking-aware negotiation. Edge AI refers to performing intelligent decision-making directly at the edge of the system (close to data sources), which supports responsive and decentralized operation without centralized model inference \cite{Mao2024Edge}. In the proposed framework, decision-making is carried out locally at the machine level with a lightweight model, without reliance on centralized coordination. The approach retains the flexibility of bidding-based mechanisms while augmenting them with a learning component that captures relative preferences among machines. Instead of learning absolute bid values, we formulate decision-making as a ranking problem over candidate machines. In addition, resource-awareness is incorporated into a negotiation process. This allows agents to implicitly account for shared resource contention while preserving decentralized operation. The proposed framework is integrated into a digital-twin system that exposes the decentralized allocation mechanism through a configurable service interface. It is important to note that while ranking-based learning has been studied in areas such as information retrieval and recommendation systems, its role in decentralized, negotiation-based task allocation remains largely underexplored. In such systems, allocation decisions are governed exclusively by the relative ordering of candidate machines. Therefore, ranking is not an alternative learning formulation. It is an aligned objective that directly determines system behavior. This work leverages this observation to redesign the learning component. Our framework is not developed as a collection of independent allocation methods. It follows a progressive construction. A resource-aware heuristic first establishes the decentralized bidding structure. An Edge-AI-based regression model then replaces handcrafted bid estimation with learned local approximation at the machine level. Finally, an autoencoder-regularized pairwise ranking model is trained using counterfactual preference pairs to modify relative bid ordering rather than approximate absolute values. Its decoder is used only during offline training. The decoding part is discarded before deployment and the compact encoder and ranking head are retained for inference. Our framework specifically targets the operational layer of circular smart manufacturing, where manufacturing and remanufacturing tasks must be allocated under contention for reusable production assets and heterogeneous machine energy characteristics. By placing lightweight ranking intelligence at machine nodes, the framework enables timely and decentralized coordination without reliance on centralized scheduling. This paper intends to make the following contributions to the field of distributed task allocation.

\begin{enumerate}
\item It proposes a decentralized task allocation framework based on negotiation among autonomous machines, designed for dynamic and resource-constrained smart manufacturing environments;

\item It introduces a progressive Edge-AI formulation that moves from local bid approximation to autoencoder-regularized pairwise ranking trained on counterfactual preferences;

\item It develops a resource-aware negotiation mechanism that incorporates shared resource contention into decentralized decision-making; and

\item It provides a paired evaluation using fully held-out simulation seeds, together with workload-sensitivity analysis, demonstrating that the benefit of learning arises from aligning the learning objective.

\end{enumerate}

Moreover, this work identifies a key limitation of existing learning-based approaches, namely the misalignment between absolute-value prediction and the inherently ranking-based nature of allocation decisions. It further provides a paired experimental evaluation across high shared-resource dependence levels and offered workloads, showing that the proposed method increases completed tasks and improves average tardiness and deadline adherence. Energy per completed task also decreases. Together, these outcomes support more effective coordination of shared production resources at the operational layer of circular smart manufacturing. Our digital twin system provides an implementation layer through which the proposed allocation mechanism can be configured, executed, and monitored. The remainder of this paper is organized as follows. Section~\ref{sec:related_work} reviews related work in decentralized task allocation and negotiation-based coordination. Section~\ref{sec:problem_formulation} presents the formal problem formulation. Section~\ref{sec:proposed_method} describes the proposed framework. Section~\ref{sec:experimental_setup} outlines the experimental setup. Section~\ref{sec:results_discussion} presents and analyzes the results. Finally, Section~\ref{sec:conclusion} concludes the paper.

\section{Related Work}\label{sec:related_work}
Task allocation and scheduling in manufacturing systems have been investigated through various paradigms, i.e., centralized optimization, distributed coordination, and learning-based decision-making \cite{Dhanaraj2025Proactive,Lan2025QoE,Li2025Dynamic}. Efficient task allocation in circular smart manufacturing requires consideration of shared-resource utilization and reuse. The energy characteristics of manufacturing operations must be taken into account. Traditional approaches typically formulate task allocation as a centralized optimization problem. A global scheduler assigns tasks to machines \cite{Li2025Decoupling}. These methods encompass mathematical programming, heuristic scheduling, and rule-based dispatch strategies \cite{Huang2024Evolutionary}. While they might yield reasonable solutions under controlled conditions, their effectiveness is constrained in modern environments due to scalability, complexity, and the need for timely responsiveness. To address the limitations of centralized approaches, distributed and multi-agent systems have been proposed. In these frameworks, task allocation is performed based on local information and improving scalability. For example, Canzini et al. have developed a multi-agent learning framework for fixture layout planning, demonstrating improved optimization performance in complex manufacturing tasks \cite{Canzini2025Decision}. Yan et al. have proposed an adaptive control scheme for multi-agent systems under event-triggered communication \cite{Yan2026Adaptive}. They focus on reducing communication overhead while maintaining system stability. Zhang et al., have investigated secure consensus in nonlinear multi-agent systems under cyberattacks and delays. They introduce event-triggered control mechanisms to enhance robustness and efficiency \cite{Zhang2025Event}. Despite these advances, most existing studies focus on control stability, optimization, or communication efficiency, and do not explicitly address how learning formulations influence decentralized decision-making and task allocation outcomes. Moreover, they may struggle to achieve globally efficient outcomes due to limited system visibility and the absence of explicit coordination among agents, particularly in the presence of shared resources or interdependent constraints.

In light of these limitations, negotiation-based mechanisms (i.e., auction-based and bidding strategies) provide a flexible and scalable approach for decentralized coordination. In such systems, agents submit bids that reflect their estimated execution cost or utility, and tasks are allocated according to a selection rule, typically based on bid minimization. Tang et al. propose a multi-agent bidding strategy within an auction-based learning framework, leveraging clustering techniques to enhance scalability and bidding efficiency \cite{Tang2024Efficient}. Similarly, Xu et al. introduce an iterative double-sided auction mechanism for data task allocation \cite{Xu2023IDADET}. Despite these advances, existing auction-based approaches primarily focus on efficiency, scalability, or convergence properties. They do not explicitly address how learning formulations influence bid ordering. They rely on handcrafted formulations that reflect local machine states, i.e., queue length or processing capability. As a result, they may fail to capture system-level interactions, especially in environments with shared resources. This limitation becomes particularly relevant in circular manufacturing settings, as efficient reuse of shared assets requires coordination beyond purely local reasoning.
More recently, learning-based approaches have been explored to improve task allocation decisions in dynamic environments \cite{Lu2026Distributed,Lv2025Deep}. These methods typically employ learning mechanisms to estimate performance metrics such as completion time, cost, or utility, which are used to guide scheduling decisions. However, a fundamental limitation of many existing learning-based approaches is their focus on predicting absolute performance values. In decentralized allocation settings, decisions are inherently comparative, as task assignment depends on the relative ordering of candidate machines. Consequently, improving prediction accuracy in absolute terms does not necessarily lead to improved allocation outcomes. In addition, most existing learning-based solutions rely on training and inference pipelines. Such designs limit scalability and introduce communication overhead, which makes them less suitable for latency-sensitive and distributed manufacturing environments. These limitations highlight a misalignment not only between learning objectives and decision-making structure, but also between system design and deployment requirements. In particular, there is a need for learning mechanisms that can capture relative decision structure and operate directly at the level of decentralized agents. 

In summary, existing studies provide important advances in decentralized scheduling, auction-based allocation, and learning-assisted decision-making. However, they rarely examine how the learning objective itself interacts with the winner-selection rule of decentralized negotiation. This gap motivates the ranking-aware formulation proposed in this paper. We explicitly target the ordering structure that determines allocation outcomes. The proposed approach integrates an Edge-AI-driven mechanism into bidding-based coordination, which enables machines to perform decision-making locally, without reliance on centralized control. By aligning the learning objective with the decision-making process, the model directly influences task allocation outcomes through relative preference modeling. In addition, the framework incorporates resource awareness into the negotiation mechanism. This mechanism supports the efficient coordination and reuse of shared resources.

\section{Problem Formulation}
\label{sec:problem_formulation}
We consider a smart system in which heterogeneous machines need to allocate and execute dynamically arriving tasks under processing, deadline, and shared resource constraints. In circular smart manufacturing settings, such constraints are further shaped by the need to efficiently utilize reusable resources and maintain energy-aware operation. The objective is to design a decentralized negotiation mechanism that enables efficient allocation while preserving scalability, responsiveness, and resource efficiency. Let $\mathcal{M} = \{m_1, m_2, \dots, m_M\}$ denote the set of machines, and $\mathcal{J} = \{j_1, j_2, \dots\}$ the set of tasks generated over time. Each task $j \in \mathcal{J}$ is characterized by the attribute vector 
\begin{equation}
\boldsymbol{\tau}_j = \left(a_j, \kappa_j, \rho_j, d_j, \pi_j, \chi_j\right),
\end{equation}
where $a_j \in \mathbb{R}_{\geq 0}$ is the arrival time, $\kappa_j \in \mathcal{K}$ is the task type, $\rho_j \in \mathbb{R}_{>0}$ is the processing requirement, $d_j \in \mathbb{R}_{>0}$ is the deadline, $\pi_j \in \Pi$ is the priority level, and $\chi_j \in \{0,1\}$ indicates whether the task requires access to a shared resource. Each machine $m \in \mathcal{M}$ is characterized by
\begin{equation}
\boldsymbol{\mu}_m = \left(\mathcal{K}_m, s_m, \varepsilon_m\right),
\end{equation}
where $\mathcal{K}_m \subseteq \mathcal{K}$ denotes the set of task types processable by machine $m$, $s_m$ is its processing speed, and $\varepsilon_m$ is its energy consumption rate. A task $j$ is compatible with machine $m$ if and only if
$\kappa_j \in \mathcal{K}_m$. Accordingly, the feasible machine set for task $j$ is defined as $ \mathcal{M}_j = \left\{ m \in \mathcal{M} \mid \kappa_j \in \mathcal{K}_m \right\}$.

Each machine maintains a local ordered queue of assigned tasks waiting for execution. Let $\mathcal{Q}_m(t)$ denote this waiting queue at time $t$; the task currently in service is not included in $\mathcal{Q}_m(t)$ and is represented separately by the busy-state indicator $\mathbb{I}_m(t)$. Task execution is assumed to be non-preemptive and sequential. The estimated processing time of task $j$ on machine $m$ is defined as
\begin{equation}
p_{j,m} = \frac{\rho_j}{s_m},
\end{equation}
capturing the dependence of execution duration on both task workload and machine capability. The corresponding queueing time of machine $m$ at time $t$ is given by
\begin{equation}
q_m(t) = \sum_{i \in \mathcal{Q}_m(t)} p_{i,m},
\end{equation}
which serves as a local estimate of accumulated workload. In addition, we define
\begin{equation}
\ell_m(t) = |\mathcal{Q}_m(t)| + \mathbb{I}_m(t),
\end{equation}
where $\ell_m(t)$ denotes the instantaneous machine load, defined as the total number of tasks either waiting in the queue or currently being processed, and $\mathbb{I}_m(t) \in \{0,1\}$ indicates whether machine $m$ is currently busy. If task $j$ is assigned to machine $m$ at time $t$, its analytical completion-time proxy is
\begin{equation}
\widehat{C}_{j,m}(t) = t + q_m(t) + p_{j,m}.
\end{equation}
Based on this proxy, the corresponding slack is defined as
\begin{equation}
s_{j,m}(t) = d_j - \widehat{C}_{j,m}(t),
\end{equation}
where negative slack indicates a potential deadline violation. Apart from the machine-local constraints, the system contains a shared resource, such as a tool or specialized device, that can be accessed by at most one task at a time. The pressure associated with this resource is represented by the conservative contention proxy, i.e.,
\begin{equation}
u(t)=\sum_{m\in\mathcal{M}}\left(
\sum_{i\in\mathcal{Q}_m(t)}\chi_i+
\mathbb{I}_m(t)\right).
\end{equation}
The inner sum counts queued tasks requiring the shared resource, and the busy-state term accounts for active execution that is not retained in the waiting queue. From a circular manufacturing perspective, the shared resource can be interpreted as a reusable production asset whose efficient utilization is critical to overall system performance. The quantity $u(t)$ acts as a compact system-level signal reflecting the current resource pressure. For tasks with $\chi_j=1$, the allocation decision should account not only for local queue conditions but also for this additional source of contention. To preserve decentralized operation while accounting for shared-resource interactions, the proposed framework uses $u(t)$ as a tractable proxy for embedding resource awareness into local decision-making.

\subsection{Decentralized Negotiation-Based Allocation}
\label{subsec:decentralized_allocation}
Task allocation is modeled as a decentralized negotiation process. Upon the arrival of a task $j$ at time $t$, each compatible machine $m \in \mathcal{M}_j$ constructs a task-machine feature vector $\mathbf{x}_{j,m}(t)$ from its local state and the shared-resource contention signal. In the proposed method, a compact learned scoring function $f\!\left(\mathbf{x}_{j,m}(t)\right)$ produces a ranking correction that is combined with the resource-aware analytical bid according to the adopted bidding formulation. The learned inference path consists only of the compact encoder and ranking head and is executed locally at each machine. Each candidate machine sends its scalar analytical bid and learned ranking score to the task-originating negotiation agent, which performs candidate-wise normalization, constructs the corrected bids, and selects the winning machine. This arrangement preserves distributed machine-level inference while coordinating winner selection separately for each arriving task, without requiring a centralized global scheduler. The feature vector is defined as

\begin{equation}
\mathbf{x}_{j,m}(t) =
\left[
\begin{aligned}
&q_m(t),\; p_{j,m},\; s_{j,m}(t),\\
&\ell_m(t),\; \chi_j,\; u(t),\\
&\pi_j,\; s_m,\; \varepsilon_m
\end{aligned}
\right]^\top,
\end{equation}
which captures both local machine state and system-level signals related to shared resource contention. The assignment decision is determined by the minimum-bid rule
\begin{equation}
\label{winner_selection}
m_j^\star(t) = \arg\min_{m \in \mathcal{M}_j} b_{j,m}(t).
\end{equation}
Thus, the allocation outcome emerges from decentralized competition among machines.

\subsection{Ranking-Based Learning Formulation}
\label{subsec:ranking_formulation}

A central observation of this work is that task allocation decisions depend on the \emph{relative ordering} of candidate machines instead of the absolute magnitude of predicted bid values. Therefore, learning absolute bid values alone is not sufficient to improve system-level allocation quality. To address this issue, we formulate machine evaluation as a ranking problem. For a given task $j$, let $\mathcal{M}_j \subseteq \mathcal{M}$ denote the set of feasible machines. For notational convenience, we represent this set as an ordered list $\{ m^{(1)}, \dots, m^{(K_j)} \}$, where $K_j = |\mathcal{M}_j|$. The objective is to learn an ordering relation over $\mathcal{M}_j$ such that more favorable machines receive lower bid values. Formally, the learned function $f(\cdot)$ is expected to induce a relative preference structure such that
\begin{equation}
b_{j,m_a}(t) < b_{j,m_b}(t)
\quad \Leftrightarrow \quad
m_a \succ_j m_b,
\end{equation}
where $m_a \succ_j m_b$ denotes that machine $m_a$ is preferred over $m_b$ for task $j$. The preference relation $m_a \succ_j m_b$ is constructed offline from counterfactual evaluations of alternative candidate-machine assignments under matched task and system conditions. These alternatives are compared by using operational outcomes that account for delay, deadline misses, completed tasks, and machine processing energy per completed task, thereby providing preference supervision without altering the negotiation mechanism. This formulation is particularly important in decentralized allocation, because only changes in relative ordering can influence assignment outcomes. As a result, ranking-aware learning directly shapes the negotiation process. It is important to note that, in decentralized auction-based allocation, the assignment decision is determined by the winner-selection rule in Eq. \eqref{winner_selection}. Therefore, any strictly increasing transformation of the bid values preserves the same allocation outcome, as long as the ordering of candidate machines remains unchanged. Consequently, only the relative ordering of bids affects system behavior. From this perspective, minimizing absolute prediction error is neither necessary nor sufficient to improve allocation performance unless it induces a more favorable ordering among candidate machines. This observation provides a formal motivation for adopting a ranking-based learning formulation.

\subsection{Performance Objectives}
\label{subsec:objectives}

To evaluate the effectiveness of the framework, we define a set of system-level performance metrics. Let $\mathcal{J}_c \subseteq \mathcal{J}$ denote the set of tasks completed within the simulation horizon. The realized completion time of task $j \in \mathcal{J}_c$ is denoted by $C_j$, and its corresponding tardiness is defined as $\Delta_j = \max(0, C_j - d_j).$ System performance is evaluated via complementary criteria that capture timeliness, deadline adherence, finite-horizon throughput, and energy efficiency. The average tardiness is defined as
\begin{equation}
\bar{\Delta}=\frac{1}{|\mathcal{J}_c|}
\sum_{j \in \mathcal{J}_c} \Delta_j,
\end{equation}
which reflects the overall timeliness of completed tasks. The number of deadline misses is given by
\begin{equation}
N_{\mathrm{miss}} =\sum_{j \in \mathcal{J}_c}
\mathbb{I}(\Delta_j > 0).
\end{equation}
To account for differences in the number of completed tasks, deadline adherence is reported through the deadline-miss rate
\begin{equation}
R_{\mathrm{miss}}=\frac{N_{\mathrm{miss}}}{|\mathcal{J}_c|}.
\end{equation}

Finite-horizon throughput is quantified by the number of completed tasks, i.e., $N_{\mathrm{comp}} = |\mathcal{J}_c|.$ When normalized results are reported, the corresponding completion rate is
\begin{equation}
R_{\mathrm{comp}} =
\frac{N_{\mathrm{comp}}}{|\mathcal{J}_a|},
\end{equation}
where $\mathcal{J}_a$ denotes the set of tasks arriving within the simulation horizon. Because tardiness and deadline-miss rate are evaluated over completed tasks, these metrics are interpreted jointly with $N_{\mathrm{comp}}$ or $R_{\mathrm{comp}}$ to avoid improvements arising from evaluating fewer completed tasks. Processing-energy efficiency is measured as the energy consumed per completed task,
\begin{equation}
E_{\mathrm{comp}} =\frac{1}{N_{\mathrm{comp}}}
\sum_{m \in \mathcal{M}}
\varepsilon_m
\sum_{j \in \mathcal{J}_m^c}
p_{j,m},
\end{equation}
where $\mathcal{J}_m^c \subseteq \mathcal{J}_c$ denotes the set of tasks completed by machine $m$. This normalization enables energy consumption to be interpreted jointly with throughput and avoids treating lower total energy caused by fewer completed tasks as an efficiency improvement.

These objectives are inherently multi-criteria and reflect trade-offs among responsiveness, deadline adherence, throughput, and energy efficiency. The role of the proposed framework is therefore not to optimize a single metric in isolation, but to produce decentralized bid rankings that improve the overall balance among these competing objectives while maintaining scalability and adaptability.

\section{Proposed Method}
\label{sec:proposed_method}

Building on the problem formulation, we introduce a decentralized task allocation framework based on negotiation among machines. The framework is developed progressively rather than as a set of independent competing methods. We first define a resource-aware heuristic bidding mechanism to establish the decentralized allocation structure. We then embed an Edge-AI-based learned decision function at the machine level to replace purely handcrafted bid estimation with local learned approximation. Finally, we redesign the learning objective as a ranking problem, so that the learned component directly shapes the bid ordering that determines decentralized task assignment. This progressive construction highlights how increasingly richer decision layers improve coordination in decentralized environments.

\begin{figure*}[t]
\centering
\includegraphics[width=\textwidth]{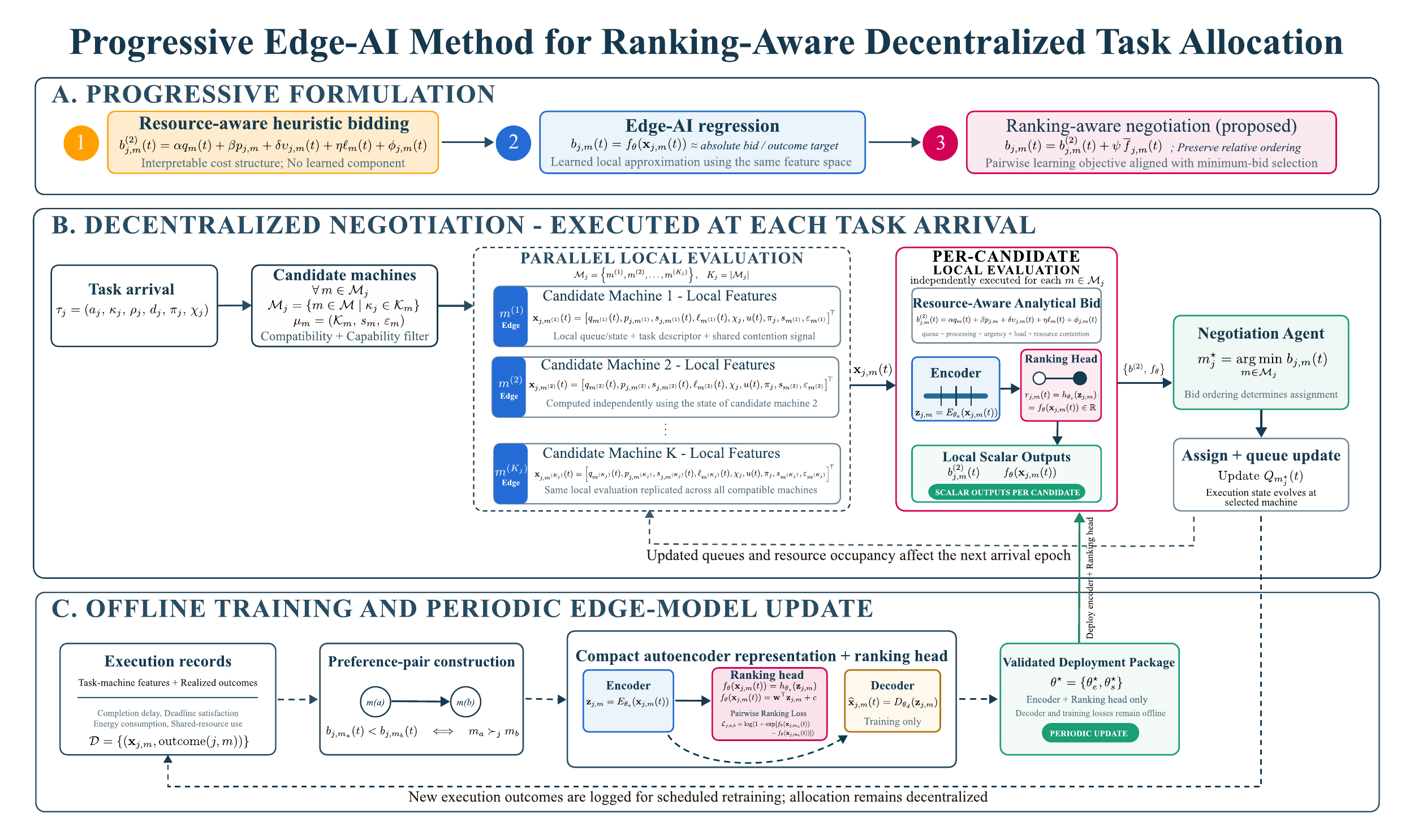}
\caption{Overview of the progressive methodological development of the Edge-AI framework. The resource-aware heuristic establishes the decentralized bidding structure, the regression formulation serves as an intermediate comparative stage for learned bid approximation, and the final proposed method combines the analytical bid with a ranking-aware correction. The lower panels illustrate decentralized online allocation, offline ranking-model training, and periodic edge-model updates.}
    \label{fig:progressive-method}
\end{figure*}

We begin with a decentralized baseline in which each machine evaluates incoming tasks using only local state information. For a task $j$ arriving at time $t$, each machine $m \in \mathcal{M}_j$ computes a bid value based on its queueing state and processing capability, i.e.,
\begin{equation}
b^{(0)}_{j,m}(t) = \alpha \, q_m(t) + \beta \, p_{j,m},
\end{equation}
where $q_m(t)$ denotes the current queueing time and $p_{j,m}$ is the processing time defined in Section~\ref{sec:problem_formulation}. The coefficients $\alpha, \beta$ determine the relative importance of waiting time and execution duration. This formulation represents a purely local estimate of execution cost. While it enables load balancing and favors faster machines, it does not account for temporal constraints or interactions across machines, and therefore may lead to suboptimal system-level behavior under constrained conditions. To improve decision quality, the bidding function is extended to incorporate additional task and machine-level features. In particular, the slack $s_{j,m}(t)$ and load $\ell_m(t)$ are included to capture temporal urgency and machine utilization. We define an urgency function based on slack, i.e.,
\begin{equation}
\upsilon_{j,m}(t) =
\begin{cases}
\gamma_1, & s_{j,m}(t) \leq 0, \\
\dfrac{1}{s_{j,m}(t)+1}, & \text{otherwise},
\end{cases}
\end{equation}
where $\gamma_1>0$ is a penalty coefficient for tasks predicted to miss their deadlines. The resulting bidding function is
\begin{equation}
b^{(1)}_{j,m}(t) =
\alpha q_m(t)+\beta p_{j,m} +
\delta \upsilon_{j,m}(t) +
\eta \ell_m(t),
\end{equation}
where $\delta$ and $\eta$ control the influence of urgency and load. Although this formulation improves responsiveness to deadlines and workload, it remains fundamentally local and does not explicitly capture interactions due to shared resources or system-level coupling. To represent shared-resource pressure explicitly within the analytical bid, we introduce a task-level contention term based on the proxy $u(t)$. This term is active for tasks requiring the shared resource ($\chi_j=1$), and is defined as
\begin{equation}
\phi_{j,m}(t)=\lambda \chi_j u(t),
\label{eq:resource_aware_penalty}
\end{equation}
where $\lambda>0$ controls the influence of shared-resource contention. The resource-aware bidding function is then given by
\begin{equation}
b^{(2)}_{j,m}(t)=\alpha q_m(t)+\beta p_{j,m}+
\delta\upsilon_{j,m}(t)+\eta\ell_m(t)+\phi_{j,m}(t).
\end{equation}

For a given task $j$, the contention term ($\phi_{j,m}(t)=\lambda\chi_j u(t)$) is identical across all feasible candidate machines. Consequently, adding this term does not change the analytical bid ordering or the minimum-bid allocation decision. The term is retained to represent shared-resource pressure explicitly in the bid formulation; no improvement in heuristic allocation is attributed to this common additive term alone. In the proposed ranking-based method, $u(t)$ is also included in the task–machine feature vector. Through its interactions with machine-specific features in the learned model, this signal can influence the ranking correction and, consequently, the final allocation decision. As an intermediate learning step, the regression model uses the same preprocessed task-machine feature vector. Its scalar output is trained by using mean-squared error to approximate the resource-aware analytical bid $b_{j,m}^{(2)}(t)$ and is directly used in the minimum-bid winner-selection rule in Eq. \eqref{winner_selection}.

\subsection{Edge-AI-Driven Ranking-Based Negotiation}
\label{subsec:edge_ai_ranking}

The key contribution of the proposed learning component is to reshape the relative ordering of machine bids rather than to approximate an absolute performance quantity. For an arriving task $j$, only the
compatible machines in $\mathcal{M}_{j}$ participate in the negotiation. Each candidate machine $m\in\mathcal{M}_{j}$ constructs the task-machine feature vector $\mathbf{x}_{j,m}(t)$ defined in
Section~\ref{subsec:decentralized_allocation} and evaluates a shared ranking function $f_{\boldsymbol{\theta}}:\mathbb{R}^{9}\rightarrow\mathbb{R}$. The resulting scalar $f_{\boldsymbol{\theta}}(\mathbf{x}_{j,m}(t))$ represents a learned ordering score, where a lower score indicates a more favorable task-machine assignment. For two candidate machines $m_a,m_b\in\mathcal{M}_{j}$, the preference relation $m_a\succ_j m_b$ indicates that $m_a$ is preferred to $m_b$ for task $j$. Consistently with the minimum-bid allocation rule, the ranking model is trained to satisfy
\begin{equation}
m_a\succ_j m_b
\quad\Longrightarrow\quad
f_{\boldsymbol{\theta}}
\left(\mathbf{x}_{j,m_a}(t)\right)
<
f_{\boldsymbol{\theta}}
\left(\mathbf{x}_{j,m_b}(t)\right).
\label{eq:desired_ranking_order}
\end{equation}

Because pairwise training determines the relative ordering of the learned scores but does not impose a fixed output scale, the scores are normalized within the feasible candidate set before being combined with the analytical bid. Let $\overline{f}_{j,m}(t)$ denote the candidate-wise centered, scale-normalized, and clipped ranking
correction defined in Algorithm~\ref{alg:edge_ai_bidding}. The final bid is then given by
\begin{equation}
b_{j,m}(t)
= b_{j,m}^{(2)}(t) + \psi\,\overline{f}_{j,m}(t),
\qquad
\psi>0,
\label{eq:ranking_corrected_bid}
\end{equation}
where $b_{j,m}^{(2)}(t)$ is the resource-aware bid defined previously and $\psi$ controls the influence of the normalized ranking correction. This formulation retains the interpretable structure of the analytical bid while allowing the learned component to modify its relative ordering. The candidate-wise normalization also removes arbitrary location and positive scale differences in the raw ranking scores before their additive combination with the analytical bid. Each candidate machine transmits its scalar analytical bid and ranking score to the task-originating agent. The agent performs the candidate-wise normalization and corrected-bid construction and then selects
\begin{equation}
m_j^{\star}(t)
=\arg\min_{m\in\mathcal{M}_{j}}b_{j,m}(t).
\label{eq:ranking_winner_selection}
\end{equation}
Therefore, the learned component affects allocation by modifying bid ordering, which directly determines the outcome of the decentralized winner-selection rule. Only the analytical bid and ranking score are communicated for each candidate; the local feature vectors, queue states, and model computations remain at the corresponding machine nodes.

\subsection{Edge-AI Model Architecture, Training, and Deployment}
\label{subsec:edge_ai_training}

The ranking function $f_{\boldsymbol{\theta}}(\cdot)$ is implemented as an autoencoder-regularized ranking model. Let $\mathcal{T}(\cdot)$ denote the feature-wise preprocessing transformation that standardizes all components of the task-machine feature vector using statistics estimated exclusively from the training subset. The fitted transformation is subsequently applied unchanged during validation, testing, and inference. Accordingly, the preprocessed input is
\begin{equation}
\widetilde{\mathbf{x}}_{j,m}(t)
=\mathcal{T}\left(\mathbf{x}_{j,m}(t)\right).
\label{eq:standardized_feature_vector}
\end{equation}

We denote $d_x$ as input and $d_z$ as latent dimensions. A shared encoder $g_{\boldsymbol{\theta}_{e}}: \mathbb{R}^{d_x}\rightarrow\mathbb{R}^{d_z}$ maps the preprocessed task-machine feature vector to a compact latent representation, i.e.,
\begin{equation}
\mathbf{z}_{j,m}(t)= g_{\boldsymbol{\theta}_{e}}
\left(\widetilde{\mathbf{x}}_{j,m}(t)\right).
\label{eq:latent_representation}
\end{equation}
The encoder depth, hidden-layer widths, and latent dimension are architectural hyperparameters that may be adapted to the input dimensionality and computational requirements of the target edge platform. For the nine-dimensional feature vector considered in this work, the reported implementation uses a $9$--$16$--$8$--$4$ encoder, corresponding to $d_x=9$ and $d_z=4$. Rectified linear unit activations are applied after the two hidden layers, while the bottleneck layer is linear. A ranking head $r_{\boldsymbol{\theta}_{r}}:\mathbb{R}^{d_z}\rightarrow\mathbb{R}$ maps the latent representation to a scalar ranking score. The ranking function is given by
\begin{equation}
f_{\boldsymbol{\theta}}
\left(\mathbf{x}_{j,m}(t)
\right)=r_{\boldsymbol{\theta}_{r}}\left(g_{\boldsymbol{\theta}_{e}}
\left(\mathcal{T}
\left(\mathbf{x}_{j,m}(t)\right)\right)\right),
\label{eq:complete_ranking_function}
\end{equation}
where $\boldsymbol{\theta}=\left\{\boldsymbol{\theta}_{e}, \boldsymbol{\theta}_{r}\right\}$ denotes the parameters required during inference. During offline training, a decoder $h_{\boldsymbol{\theta}_{d}}:\mathbb{R}^{d_z}\rightarrow\mathbb{R}^{d_x}$ reconstructs the preprocessed feature vector from its latent representation, i.e.,
\begin{equation}
\widehat{\mathbf{x}}_{j,m}(t)=
h_{\boldsymbol{\theta}_{d}}\left(\mathbf{z}_{j,m}(t)\right).
\label{eq:feature_reconstruction}
\end{equation}
The decoder architecture can be selected according to the encoder
configuration and is used only to regularize the latent representation
during training. In the reported implementation, a symmetric
$4$--$8$--$16$--$9$ decoder is employed, with rectified linear unit
activations in its hidden layers and a linear output layer. The full
set of parameters used during training is $\boldsymbol{\Theta}=\left\{\boldsymbol{\theta}_{e},\boldsymbol{\theta}_{r},\boldsymbol{\theta}_{d}\right\}.$ The number of parameters depends on the selected encoder and decoder configurations and is therefore not an intrinsic restriction of the proposed formulation. For the architecture instantiated in this work, the complete training network contains 674 trainable parameters, i.e., 332 in the encoder, 5 in the ranking head, and 337 in the decoder. After offline training, the decoder is discarded, leaving only the encoder and ranking head, with 337 parameters, for online inference. These parameters require approximately $1.32$~KiB of raw FP32 weight storage. This value excludes preprocessing statistics, intermediate activations, and runtime or framework overhead.

\subsubsection{Pairwise Preferences}
The proposed ranking formulation requires an ordered preference relation between alternative candidate machines. In general, the preference pairs may be constructed using any consistent operational criterion capable of comparing alternative task--machine assignments. In this work, the preferences are generated offline through counterfactual simulation rollouts. For each recorded allocation state, the corresponding task is assigned to each feasible machine in a separate rollout, while the scenario configuration, random seed, and system conditions preceding the allocation decision are held fixed. Before each rollout, the complete pre-allocation simulator state and random-generator state are restored. After the assignment, subsequent tasks are handled by using the same fixed continuation policy. The restored random state ensures identical future task arrivals, task attributes, and shared-resource requirements across all counterfactual branches. Each forced assignment produces a system-level outcome that reflects the downstream effect of selecting the corresponding candidate machine. The particular outcome criteria and their ordering are design choices that may be adapted to the operational objectives of the considered system. In the reported implementation, candidate outcomes are compared lexicographically using completed tasks, deadline-miss rate, average tardiness, and machine processing energy per completed task. A candidate producing more completed tasks is preferred first; ties are resolved sequentially by lower deadline-miss rate, lower average tardiness, and lower energy per completed task. This ordering prioritizes finite-horizon throughput so that an alternative is not favored by completing fewer tasks. The remaining criteria are applied sequentially to reflect deadline adherence, delay severity, and processing-energy efficiency without introducing arbitrary weights. Candidate assignments that remain indistinguishable under all criteria are excluded. Each resulting ordered preference pair is represented as $\left(\mathbf{x}_{i}^{+},\mathbf{x}_{i}^{-}\right),$
where $\mathbf{x}_{i}^{+}$ and $\mathbf{x}_{i}^{-}$ correspond to the
task--machine feature vectors of the preferred and non-preferred
machines, respectively. Thus, for preference pair $i$ associated with
task $j_i$,
\begin{equation}
m_i^{+}\succ_{j_i}m_i^{-}.
\end{equation}

The preference pairs are divided into training, validation, and test subsets at the simulation-episode level. All task instances and preference pairs associated with the same scenario and simulation seed are retained in the same subset. This partitioning prevents closely related system trajectories from appearing across different subsets and thereby reduces information leakage. Approximately $70\%$ of the available simulation episodes are assigned to training, $15\%$ to validation, and $15\%$ to test subsets. The preprocessing transformation $\mathcal{T}(\cdot)$ is fitted using the training subset only and is subsequently applied unchanged to the validation and test subsets.

\subsubsection{Joint Ranking and Reconstruction Objective}
For a mini-batch
\begin{equation}
\mathcal{B}=\left\{\left(
\mathbf{x}_{i}^{+},
\mathbf{x}_{i}^{-}\right)
\right\}_{i=1}^{|\mathcal{B}|},
\end{equation}
the two members of each pair are processed by the same encoder and
ranking head. The pairwise ranking loss is
\begin{equation}
\mathcal{L}_{\dagger}
=\frac{1}{|\mathcal{B}|}
\sum_{i=1}^{|\mathcal{B}|}
\log
\left[1+\exp\left(f_{\boldsymbol{\theta}}
\left(\mathbf{x}_{i}^{+}\right)
-f_{\boldsymbol{\theta}}
\left(\mathbf{x}_{i}^{-}
\right)\right)\right].
\label{eq:batch_pairwise_loss}
\end{equation}
Because the preferred candidate is placed first in each ordered pair,
minimizing Eq. \eqref{eq:batch_pairwise_loss} encourages
\begin{equation}
f_{\boldsymbol{\theta}}
\left(\mathbf{x}_{i}^{+}
\right)<f_{\boldsymbol{\theta}}
\left(\mathbf{x}_{i}^{-}\right).
\end{equation}
Accordingly, preferred machines receive lower ranking scores, given the minimum-bid winner-selection rule. For compactness, we define $
\widetilde{\mathbf{x}}_{i}^{+}=\mathcal{T}
\left(\mathbf{x}_{i}^{+}\right),
\widetilde{\mathbf{x}}_{i}^{-}=\mathcal{T}
\left(\mathbf{x}_{i}^{-}\right).$ The reconstruction loss includes both members of every preference pair, i.e.,
\begin{align}
\mathcal{L}_{\ddagger}=\frac{1}{2|\mathcal{B}|d_x}\sum_{i=1}^{|\mathcal{B}|}
\Bigg[&
\left\|\widetilde{\mathbf{x}}_{i}^{+}
-h_{\boldsymbol{\theta}_{d}}
\left(g_{\boldsymbol{\theta}_{e}}
\left(\widetilde{\mathbf{x}}_{i}^{+}
\right)\right)\right\|_{2}^{2}\nonumber\\+
&\left\|\widetilde{\mathbf{x}}_{i}^{-}-h_{\boldsymbol{\theta}_{d}}
\left(g_{\boldsymbol{\theta}_{e}}
\left(\widetilde{\mathbf{x}}_{i}^{-}\right)
\right)\right\|_{2}^{2}\Bigg],
\label{eq:reconstruction_loss}
\end{align}
where $d_x$ denotes the feature-vector dimension. The normalization by
$d_x$ expresses the reconstruction term as a mean squared error over
both the samples and feature dimensions. The complete training objective is
\begin{equation}
\mathcal{L}_{*}=\mathcal{L}_{\dagger}+\omega_{\ddagger}\,\mathcal{L}_{\ddagger},
\label{eq:complete_loss}
\end{equation}
where $\omega_{\ddagger}\geq 0$ controls the contribution of
reconstruction regularization. The ranking loss updates the encoder
and ranking head, whereas the reconstruction loss updates the encoder
and decoder. Consequently, the shared encoder receives gradients from
both objectives:
\begin{equation}
\nabla_{\boldsymbol{\theta}_{e}}
\mathcal{L}_{*}=\nabla_{\boldsymbol{\theta}_{e}}
\mathcal{L}_{\dagger}+\omega_{\ddagger}\nabla_{\boldsymbol{\theta}_{e}}
\mathcal{L}_{\ddagger}.
\label{eq:encoder_joint_gradient}
\end{equation}

The optimization method, learning rate, mini-batch size, regularization strength, and stopping criterion are training hyperparameters that can be selected according to the available data and computational requirements. In our reported implementation, $\omega_{\ddagger}=0.1$, and the model is trained using Adam with a learning rate of $10^{-3}$, a mini-batch size of 64, weight decay of $10^{-5}$, and a maximum of 100 epochs. Early stopping is applied when the validation value of $\mathcal{L}_{*}$ does not improve for 10 consecutive epochs. The parameter state yielding the minimum
validation loss is restored and evaluated once on the held-out pairwise test subset. Algorithm~\ref{alg:ranking_model_training} summarizes the offline training procedure. The same procedure can be repeated offline when a scheduled model update is required. The decoder, reconstruction operations, pairwise-loss evaluation, and parameter optimization remain outside the online allocation loop. Only the fitted preprocessing transformation, trained encoder, and ranking head are deployed at each machine node.

\begin{algorithm}[t]
\caption{Training/Updating the Ranking Model}
\label{alg:ranking_model_training}
\begin{algorithmic}[1]
\Require Episode-grouped task--machine records $\mathcal{D}$;
encoder $g_{\boldsymbol{\theta}_{e}}$;
decoder $h_{\boldsymbol{\theta}_{d}}$;
ranking head $r_{\boldsymbol{\theta}_{r}}$;
reconstruction weight $\omega_{\ddagger}$;
mini-batch size $B$;
maximum epochs $E$;
early-stopping patience $P$
\Ensure Deployment parameters
$\boldsymbol{\theta}^{\star}
=\{\boldsymbol{\theta}_{e}^{\star},
\boldsymbol{\theta}_{r}^{\star}\}$
and fitted preprocessing transformation $\mathcal{T}(\cdot)$

\State Partition the simulation episodes in $\mathcal{D}$ into
training, validation, and test subsets

\State Construct the counterfactual ordered preference-pair sets
$\mathcal{P}_{\mathrm{tr}}$,
$\mathcal{P}_{\mathrm{val}}$, and
$\mathcal{P}_{\mathrm{te}}$

\State Fit $\mathcal{T}(\cdot)$ using
$\mathcal{P}_{\mathrm{tr}}$ only

\State Initialize
$\boldsymbol{\Theta}
=
\{
\boldsymbol{\theta}_{e},
\boldsymbol{\theta}_{r},
\boldsymbol{\theta}_{d}
\}$

\For{$e=1,\ldots,E$}
    \For{each mini-batch
    $\mathcal{B}
    =
    \{(\mathbf{x}_{i}^{+},\mathbf{x}_{i}^{-})\}_{i=1}^{B}$
    sampled from $\mathcal{P}_{\mathrm{tr}}$}

        \State
        $\widetilde{\mathbf{x}}_{i}^{+}
        \gets
        \mathcal{T}(\mathbf{x}_{i}^{+})$
        \State
        $\widetilde{\mathbf{x}}_{i}^{-}
        \gets
        \mathcal{T}(\mathbf{x}_{i}^{-})$

        \State
        $\mathbf{z}_{i}^{+}
        \gets
        g_{\boldsymbol{\theta}_{e}}
        (\widetilde{\mathbf{x}}_{i}^{+})$
        \State
        $\mathbf{z}_{i}^{-}
        \gets
        g_{\boldsymbol{\theta}_{e}}
        (\widetilde{\mathbf{x}}_{i}^{-})$

        \State
        $\xi_{i}^{+}
        \gets
        r_{\boldsymbol{\theta}_{r}}(\mathbf{z}_{i}^{+})$
        \State
        $\xi_{i}^{-}
        \gets
        r_{\boldsymbol{\theta}_{r}}(\mathbf{z}_{i}^{-})$

        \State
        $\widehat{\mathbf{x}}_{i}^{+}
        \gets
        h_{\boldsymbol{\theta}_{d}}(\mathbf{z}_{i}^{+})$
        \State
        $\widehat{\mathbf{x}}_{i}^{-}
        \gets
        h_{\boldsymbol{\theta}_{d}}(\mathbf{z}_{i}^{-})$

        \State Compute $\mathcal{L}_{\dagger}$
        using \eqref{eq:batch_pairwise_loss}

        \State Compute $\mathcal{L}_{\ddagger}$
        using \eqref{eq:reconstruction_loss}

        \State Compute $\mathcal{L}_{*}$
        using \eqref{eq:complete_loss}

        \State Update $\boldsymbol{\Theta}$ using Adam
    \EndFor

    \State Evaluate $\mathcal{L}_{*}$ on
    $\mathcal{P}_{\mathrm{val}}$

    \State Retain the parameter state if the validation loss improves

    \If{validation $\mathcal{L}_{*}$ has not improved for $P$ epochs}
        \State \textbf{break}
    \EndIf
\EndFor

\State Restore the parameter state with minimum validation loss

\State Evaluate the selected model once on
$\mathcal{P}_{\mathrm{te}}$

\State Discard decoder $h_{\boldsymbol{\theta}_{d}}$

\State Deploy
$g_{\boldsymbol{\theta}_{e}^{\star}}$,
$r_{\boldsymbol{\theta}_{r}^{\star}}$, and
$\mathcal{T}(\cdot)$ to the machine-level edge nodes

\State \Return
$\boldsymbol{\theta}^{\star}
=
\{
\boldsymbol{\theta}_{e}^{\star},
\boldsymbol{\theta}_{r}^{\star}
\}$
and $\mathcal{T}(\cdot)$
\end{algorithmic}
\end{algorithm}

To examine the practical deployability of the proposed architecture, the online decision layer is implemented on a resource-constrained edge-computing platform. The deployed component contains only the fitted preprocessing transformation $\mathcal{T}(\cdot)$, the trained encoder $g_{\boldsymbol{\theta}_{e}^{\star}}(\cdot)$, and the ranking head $r_{\boldsymbol{\theta}_{r}^{\star}}(\cdot)$. Model training, counterfactual rollout generation, reconstruction, and parameter optimization remain offline. This separation preserves decentralized online execution and reduces the deployed model size relative to the complete training architecture. For the architecture instantiated in this work, the deployed ranking model contains 337 parameters, corresponding to approximately $1.32$~KiB of raw FP32 weight storage, excluding preprocessing statistics, intermediate activations, and runtime or framework overhead. Algorithm~\ref{alg:edge_ai_bidding} summarizes the online ranking-based allocation procedure.

\begin{algorithm}[!t]
\caption{Edge-AI-Driven Ranking-Based Task Allocation}
\label{alg:edge_ai_bidding}
\begin{algorithmic}[1]
\Require Task $j$;\
arrival time $t$;\
machines $\mathcal{M}$;\
shared-resource contention signal $u(t)$;\
preprocessing transformation $\mathcal{T}(\cdot)$;\
trained encoder $g_{\boldsymbol{\theta}_{e}^{\star}}$;\
trained ranking head $r_{\boldsymbol{\theta}_{r}^{\star}}$;\
bidding parameters $\alpha,\beta,\gamma_{1},\delta,\eta,\lambda,\psi$;\
normalization tolerance $\epsilon$;\
score-clipping threshold $c$
\Ensure Selected machine $m_j^{\star}(t)$
\State Determine
$\mathcal{M}_{j}=\{m\in\mathcal{M}\mid\kappa_j\in\mathcal{K}_{m}\}$

\For{each $m\in\mathcal{M}_{j}$ \textbf{in parallel}}
\State Compute $q_m(t)$ and $p_{j,m}$
\State Compute the analytical proxy
$\widehat{C}_{j,m}(t)$
\State Compute $s_{j,m}(t)$ and $\ell_m(t)$
\State Compute $\upsilon_{j,m}(t)$

\State
$\phi_{j,m}(t)
\gets
\lambda\chi_j u(t)$

\State Form the task-machine feature vector
$\mathbf{x}_{j,m}(t)$

\State
$\widetilde{\mathbf{x}}_{j,m}(t)
\gets\mathcal{T}\left(\mathbf{x}_{j,m}(t)\right)$

\State
$\mathbf{z}_{j,m}(t)\gets g_{\boldsymbol{\theta}_{e}^{\star}}
\left(\widetilde{\mathbf{x}}_{j,m}(t)\right)$

\State
$f_{\boldsymbol{\theta}^{\star}}
\left(\mathbf{x}_{j,m}(t)\right)
\gets r_{\boldsymbol{\theta}_{r}^{\star}}
\left(\mathbf{z}_{j,m}(t)\right)$
\State Compute the resource-aware analytical bid
$b_{j,m}^{(2)}(t)$
\State Send the scalar analytical bid
$b_{j,m}^{(2)}(t)$ and ranking score
$f_{\boldsymbol{\theta}^{\star}}
\left(\mathbf{x}_{j,m}(t)\right)$
to the task-originating agent
\EndFor
\State At the task-originating agent, compute the
candidate-wise mean ranking score
\begin{equation*}
\mu_j\gets\frac{1}{|\mathcal{M}_j|}\sum_{m\in\mathcal{M}_j}f_{\boldsymbol{\theta}^{\star}}
\left(\mathbf{x}_{j,m}(t)\right)
\end{equation*}
\State At the task-originating agent, compute the
candidate-wise score scale
\begin{equation*}
\sigma_j
\gets\sqrt{\frac{1}{|\mathcal{M}_j|}\sum_{m\in\mathcal{M}_j}
\left[f_{\boldsymbol{\theta}^{\star}}\left(\mathbf{x}_{j,m}(t)\right)-\mu_j\right]^2}
\end{equation*}
\State At the task-originating agent, construct the
normalized corrections and corrected bids
\For{each $m\in\mathcal{M}_{j}$}
\If{$\sigma_j\leq\epsilon$}
\State
$\overline{f}_{j,m}(t)\gets0$
\Else
\State
$\overline{f}_{j,m}(t)
\gets
\operatorname{clip}
\left(\dfrac{f_{\boldsymbol{\theta}^{\star}}
\left(\mathbf{x}_{j,m}(t)\right)
-\mu_j}{\sigma_j},-c,c\right)$
\EndIf
\State
$b_{j,m}(t)
\gets
b_{j,m}^{(2)}(t)+\psi\,\overline{f}_{j,m}(t)$
\EndFor

\State
$m_j^{\star}(t)
\gets\arg\min_{m\in\mathcal{M}_{j}}b_{j,m}(t)$
\State Assign task $j$ to $m_j^{\star}(t)$
\State Update
$\mathcal{Q}_{m_j^{\star}(t)}(t)$

\State Register task $j$ for execution-record logging upon completion
\State \Return $m_j^{\star}(t)$
\end{algorithmic}
\end{algorithm}

\section{Experimental Setup}
\label{sec:experimental_setup}

The experimental evaluation is designed to examine how different learning formulations affect decentralized task allocation under dynamic workloads, temporal constraints, and shared-resource contention. A discrete-event simulation is used to model the manufacturing environment as a stochastic and event-driven system. This enables us to systematically evaluate the allocation mechanism under controlled and reproducible conditions. The system consists of the heterogeneous machine set $\mathcal{M}$ defined in Section~\ref{sec:problem_formulation}. Tasks are allocated online upon arrival, enter machine-local queues, and are processed non-preemptively. Each compatible machine evaluates the arriving task using its local state and the shared-resource contention signal. The task-originating negotiation agent then performs the lightweight score aggregation and winner selection required for allocation, without centralized model inference or global scheduling. A unit-capacity shared resource introduces coupling among otherwise decentralized machines. Its occupancy evolves according to task execution, thereby capturing resource contention and non-local dependencies. From the standpoint of circular manufacturing, the shared resource represents a reusable production capability whose coordinated utilization affects both operational efficiency and capacity utilization. The evaluation is structured in accordance with the progressive development of the proposed framework. We first consider the resource-aware heuristic formulation, followed by a regression-based Edge-AI formulation using the same local feature space. The proposed formulation subsequently uses counterfactual pairwise preference learning to produce a ranking correction for the resource-aware analytical bids. This aligns the learned component with the ordering-based winner-selection rule. This progression enables the effect of the learning formulation to be examined while maintaining the same decentralized allocation architecture and experimental conditions. It is worth noting that reinforcement-learning, multi-agent-learning, and other Edge-AI scheduling approaches generally rely on different state and action spaces, reward functions, communication structures, and training assumptions. Adapting them to the present minimum-bid negotiation architecture would require substantial methodological redesign and would prevent a controlled comparison. Therefore, we compare the progressive formulations under the same information structure, allocation rule, system model, and workload realizations. The principal experimental configuration is summarized in Table~\ref{settings}. The analytical-bid scaling coefficients have been calibrated exclusively. We have used the validation episodes and they have been fixed across all methods, workloads, shared-resource conditions, and evaluation seeds. Their numerical magnitudes are implementation-specific because they depend on the units and operating ranges of the instantiated system.

\begin{table}[!b]
\caption{Experimental configuration}
\label{settings}
\centering
\footnotesize
\setlength{\tabcolsep}{3.5pt}
\renewcommand{\arraystretch}{1.08}
\begin{tabular}{@{}p{0.31\columnwidth}p{0.63\columnwidth}@{}}
\hline
\textbf{Configuration aspect} & \textbf{Experimental setting} \\
\hline
Evaluation coverage &
Finite horizon of 100 time units; offered workloads of
$\{35,45,55,65\}$ tasks \\
Temporal workload &
Maximum inter-arrival time $\overline{\Delta}_{a}=2$;
relative deadlines sampled from
$\mathcal{U}_{d}\{10,\ldots,15\}$ \\
Resource dependence &
Shared-resource requirement probabilities
$p_{r}\in\{0.7,0.9\}$ \\
Counterfactual corpus &
100 training, 22 validation, and 22 test episodes;
2,125 ordered training pairs \\
Held-out evaluation &
30 fully held-out evaluation seeds \\
Ranking configuration &
Ranking-correction coefficient $\psi=20.0$ and
reconstruction weight $\omega^{\ddagger}=0.1$ \\
Optimization &
Adam optimizer; learning rate $10^{-3}$; batch size 64;
weight decay $10^{-5}$ \\
Model selection &
Maximum 100 epochs; early-stopping patience of 10 epochs;
minimum-validation-loss checkpoint restored \\
Edge inference profile &
9--16--8--4 encoder and linear 4--1 ranking head;
337 deployed parameters and 1.32 KiB FP32 weight storage \\
\hline
\end{tabular}
\end{table}

\subsection{Workload Generation Model}
\label{subsec:workload_generation}

The workload is generated as a stochastic process that captures variability in arrival patterns, task characteristics, and shared-resource requirements. The arrival process is modeled as a renewal process with independent inter-arrival times, i.e.,
\begin{equation}
\Delta a_j = a_{j+1}-a_j
\sim
\mathcal{U}_{\mathrm{d}}
\left\{\Delta a_{\min},\ldots,\overline{\Delta}_a
\right\},
\end{equation}
where $\mathcal{U}_{\mathrm{d}}\{\cdot\}$ denotes a discrete uniform distribution, $a_j$ denotes the arrival time of task $j$, and $\overline{\Delta}_a$ controls the maximum inter-arrival time and, consequently, the workload intensity. Smaller values of $\overline{\Delta}_a$ produce more frequent task arrivals and higher system load. Each task $j$ is associated with the attribute vector $\boldsymbol{\tau}_j=(a_j,\kappa_j,\rho_j,d_j,\pi_j,\chi_j)$, as defined in Section~\ref{sec:problem_formulation}. The task type $\kappa_j$ is sampled from a discrete distribution over $\mathcal{K}$, thereby inducing heterogeneity in machine-compatibility constraints. The processing requirement is generated according to $\rho_j\sim\mathcal{U}_{\mathrm{d}}
\left\{\rho_{\min},\ldots,\rho_{\max} \right\},$ which produces variability in execution times across tasks and heterogeneous machines. Rather than coupling the deadline directly to the processing requirement, the relative deadline is sampled from a bounded discrete distribution, i.e., $d_j-a_j\sim\mathcal{U}_{\mathrm{d}}\left\{D_{\min},\ldots,D_{\max}\right\}$.
The bounds $D_{\min}$ and $D_{\max}$ control the available temporal
slack and can be adjusted to represent operating conditions
with different levels of deadline pressure. Task priority $\pi_j$ is sampled from a discrete set, introducing heterogeneity in task importance and providing an additional input to the machine-level decision model. The shared-resource requirement is modeled as a Bernoulli random variable, i.e., $\chi_j\sim\mathrm{Bernoulli}\!\left(p_r\right),$ where $p_r$ is the probability that an arriving task requires access to the capacity-constrained shared resource. Increasing $p_r$ increases shared-resource dependence, inter-machine coupling, and the potential for resource contention. In circular settings, this represents production environments in which a larger fraction of tasks depends on the same reusable production asset. The combination of these stochastic elements yields a workload process with controlled variability and tunable difficulty. In our experiments, the offered workload and $p_r$ are varied to evaluate performance under different levels of system load and shared-resource dependence, while the remaining workload-generation settings are held fixed. The specific values used in the experiments are reported in Section~\ref{sec:results_discussion}. The coefficients $\alpha$, $\beta$, $\delta$, $\eta$, $\lambda$, $\psi$, and $\gamma_1$ are fixed weighting hyperparameters rather than learned model parameters or physical system constants. They balance bid components with different numerical scales and are selected using validation-only calibration before the held-out evaluation. The same configuration is then retained across all methods, random seeds, offered workloads, and shared-resource settings, without scenario-specific retuning.

\subsection{Digital-Twin and Edge Deployment Prototype}
\label{subsec:digital_twin}

\begin{figure}[t]
    \centering
    \includegraphics[width=\columnwidth]{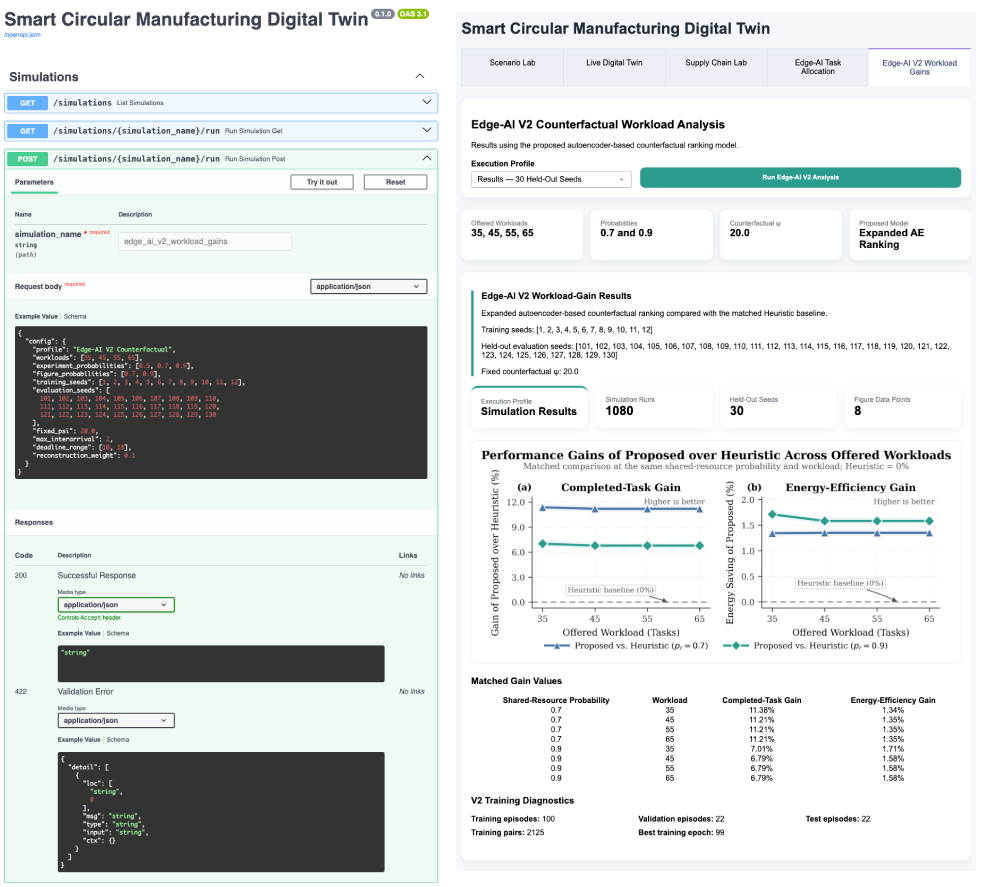}
    \caption{Python-based end-to-end DT platform developed for smart and circular manufacturing systems, with the proposed decentralized task-allocation framework integrated as a dedicated module.}
    \label{fig:digital_twin}
\end{figure}

Our framework has been integrated into a digital twin (DT) system that provides an execution and monitoring layer for the decentralized allocation mechanism. Fig.~\ref{fig:digital_twin} shows the implemented service interface and its configurable simulation endpoints. It exposes the manufacturing environment through a REST-based service interface, which allows us to configure workload, shared-resource requirements, operating conditions, and random seeds at run time. Each simulation request initializes the corresponding system configuration and invokes the proposed ranking-based task-allocation mechanism, while the resulting machine-level decisions and system-level performance indicators are returned through the same interface. The implementation separates the digital twin service layer from the machine-level Edge-AI decision logic. Model training and parameter optimization are performed offline, whereas online machine-level inference requires only the fitted preprocessing transformation, trained encoder, and ranking head described in Section~\ref{subsec:edge_ai_training}. This lightweight inference pipeline is deployed on a Raspberry Pi~4 Model~B, equipped with a quad-core 64-bit Broadcom BCM2711 Cortex-A72 processor operating at 1.8\,GHz and 4\,GB of LPDDR4 memory. Each compatible machine locally computes its scalar analytical bid and ranking score. The task-originating negotiation agent performs the lightweight candidate-wise score aggregation, corrected-bid construction, and winner selection described in Section~\ref{sec:proposed_method}. The required normalization statistics are formed at the negotiation layer using only compact candidate-level scalar outputs; no raw feature vectors, queue states, execution histories, or model data are transferred. The resulting architecture preserves distributed machine-level inference without requiring centralized model execution or global scheduling, while allowing the manufacturing state and allocation outcomes to be configured, executed, and monitored through our DT interface.

\section{Experimental Results}
\label{sec:results_discussion}

We now evaluate the proposed allocation method under high shared-resource dependence. The analysis first examines primary performance and the effect of the learning formulation, and then investigates distributional behavior, paired seed-wise changes, and sensitivity to the offered workload. The primary comparisons use an offered workload of 45 tasks and $30$ fully held-out simulation seeds. Because the compared methods may complete different numbers of tasks within the finite simulation horizon, energy efficiency is reported as energy consumption per completed task, i.e.,
\begin{equation}
E_{\mathrm{comp}}=\frac{E_{\mathrm{T}}}{N_{\mathrm{comp}}},
\label{eq:energy_per_completed_task}
\end{equation}
where $E_{\mathrm{T}}$ is the total processing energy attributable to tasks completed and $N_{\mathrm{comp}}$ is the number of those completed tasks. Processing energy consumed by tasks that remain unfinished at the end of the horizon is excluded. This normalization prevents a method from appearing energy efficient merely because it completes fewer tasks.

\subsection{Effect of Learning Formulation}
\label{subsec:primary_results}

Table~\ref{tab:primary_results} reports the primary results under shared-resource requirement probabilities $p_r=0.7$ and $p_r=0.9$. These conditions represent operating regimes in which a large fraction of arriving tasks depends on the capacity-constrained shared resource. The table compares the resource-aware heuristic, the regression-based Edge-AI baseline, and the proposed ranking-based method under identical seed-matched task streams. At $p_r=0.7$, the proposed method (Edge-AI Ranking) increases the mean number of completed tasks from $19.03$ to $21.17$, while reducing average tardiness from $22.15$ to $19.44$ and the deadline-miss rate from $0.79$ to $0.74$. At $p_r=0.9$, it increases completed tasks from $14.73$ to $15.73$, reduces average tardiness from $25.42$ to $23.17$, and lowers the deadline-miss rate from $0.83$ to $0.80$. The paired differences in these three metrics are statistically significant at both shared-resource probabilities. Thus, the reductions in tardiness and deadline-miss rate are achieved together with, rather than at the expense of, a higher number of completed tasks. The regression-based method remains close to the heuristic across the reported metrics. This result indicates that approximating the analytical bid more accurately does not necessarily produce a meaningful change in the ordering of candidate machines. In contrast, the proposed method modifies this ordering through counterfactual pairwise preference learning and consequently changes the allocation outcomes. The comparison therefore supports the motivation for aligning the learning objective with the minimum-bid winner-selection rule. It should, however, be interpreted as a comparison of learning formulations rather than as a complete architectural ablation of every component of the proposed network. The proposed method also yields the lowest sample mean for energy per completed task at both $p_r=0.7$ and $p_r=0.9$.

\begin{table*}[t]
\centering
\footnotesize
\caption{Primary performance under high shared-resource dependence at a workload of 45 tasks. Values are reported as mean $\pm$ standard deviation over $30$ fully held-out simulation seeds. Higher is better for completed tasks, whereas lower is better for all other metrics. Average tardiness is reported in simulation time units; completed tasks are counts; deadline-miss rate is dimensionless; and processing energy is expressed in simulation energy units per completed task.}
\label{tab:primary_results}
\begin{threeparttable}
\setlength{\tabcolsep}{4pt}
\renewcommand{\arraystretch}{0.92}
\begin{tabular*}{\textwidth}{
    @{\extracolsep{\fill}}
    c
    l
    c
    c
    c
    c
    @{}
}
\toprule
\(\boldsymbol{p_r}\)
&\textbf{Method}
&\textbf{Completed Tasks}
&\textbf{Avg. Tardiness}
&\textbf{Deadline-Miss Rate}
&\textbf{Processing Energy/Completed Task}
\\
\midrule
\multirow{3}{*}{0.7}
&
Heuristic
&\(19.03 \pm 2.85\)&\(22.15 \pm 4.38\)&\(0.79 \pm 0.09\)&\(14.97 \pm 0.94\)
\\
&
Regression&
\(19.07 \pm 2.88\)&\(22.32 \pm 4.34\)&\(0.78 \pm 0.09\)&\(14.88 \pm 0.94\)
\\
&
Proposed&\(\mathbf{21.17 \pm 2.63}^{\S}\)&\(\mathbf{19.44 \pm 5.78}^{\S}\)&
\(\mathbf{0.74 \pm 0.14}^{\S}\)
&
\(\mathbf{14.76 \pm 1.09}\)
\\
\midrule
\multirow{3}{*}{0.9}
&Heuristic&\(14.73 \pm 1.55\)&\(25.42 \pm 3.03\)&\(0.83 \pm 0.06\)&\(14.89 \pm 1.09\)
\\

&
Regression&\(14.70 \pm 1.53\)&\(25.38 \pm 3.08\)&\(0.83 \pm 0.07\)&\(14.92 \pm 1.08\)
\\

&
Proposed&\(\mathbf{15.73 \pm 1.66}^{\S}\)
&\(\mathbf{23.17 \pm 3.79}^{\S}\)&\(\mathbf{0.80 \pm 0.08}^{\S}\)&
\(\mathbf{14.65 \pm 1.09}\)
\\

\bottomrule
\end{tabular*}

\begin{tablenotes}[flushleft]
\footnotesize
\item
\(p_r\) denotes the probability that a task requires the limited
shared resource. The symbol \(\S\) denotes a statistically significant difference between Proposed and Heuristic on the same held-out seeds
(\(p<0.05\)).

\end{tablenotes}
\end{threeparttable}
\end{table*}

\subsection{Distributional and Seed-Wise Analysis}
\label{subsec:distribution_seedwise_results}

\begin{figure}[t]
\centering
\includegraphics[width=0.98\columnwidth]{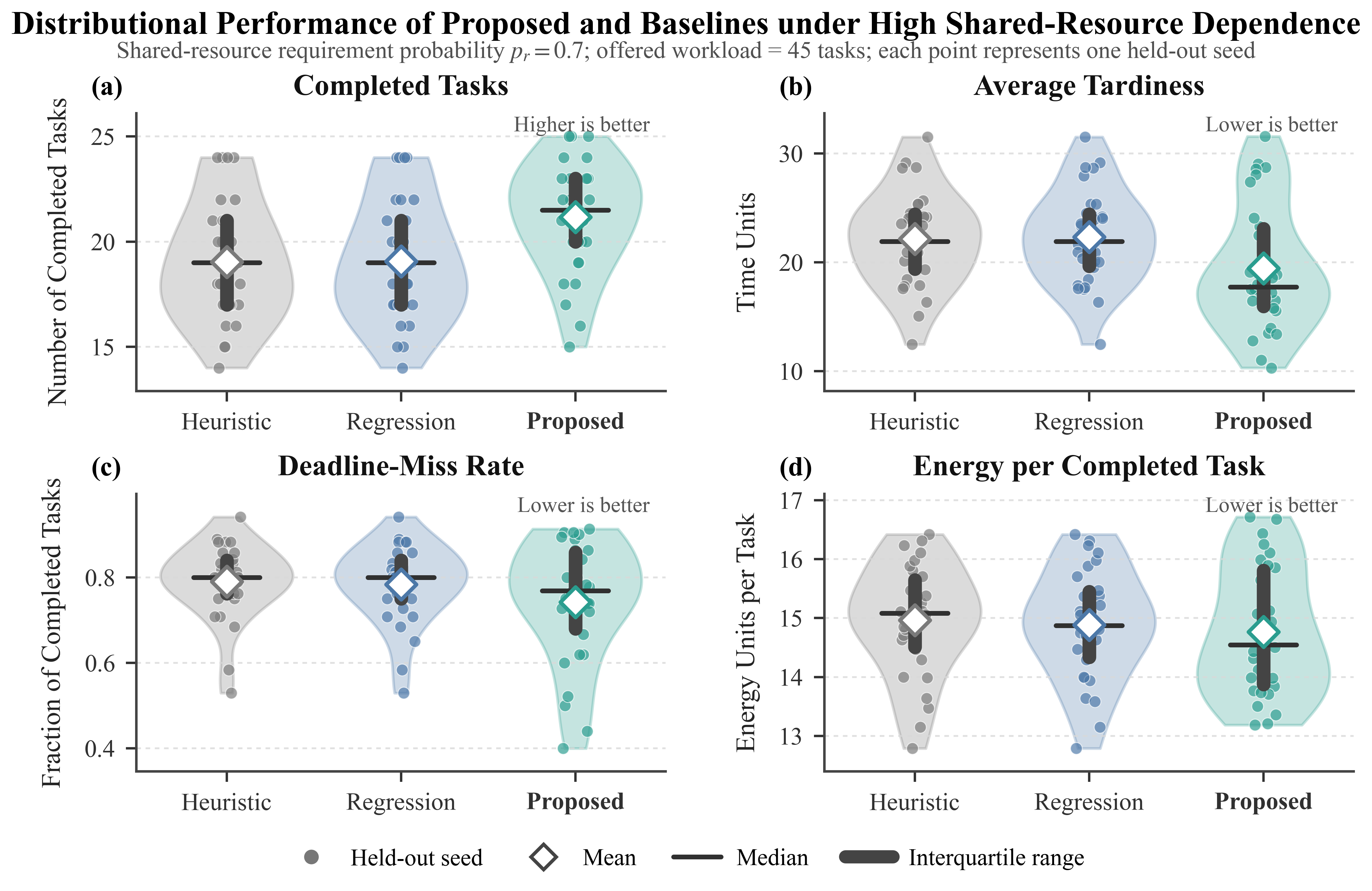}
\caption{Distributional performance at $p_r=0.7$ given 45 tasks: (a) completed tasks, (b) average tardiness, (c) deadline-miss rate, and (d) machine processing energy per completed task. Each point represents one fully held-out simulation seed. The diamond denotes the mean, the horizontal line denotes the median, and the thick vertical segment denotes the interquartile range.}
\label{fig:distributional_results}
\end{figure}

\begin{figure}[t]
\centering
\includegraphics[width=0.98\columnwidth]{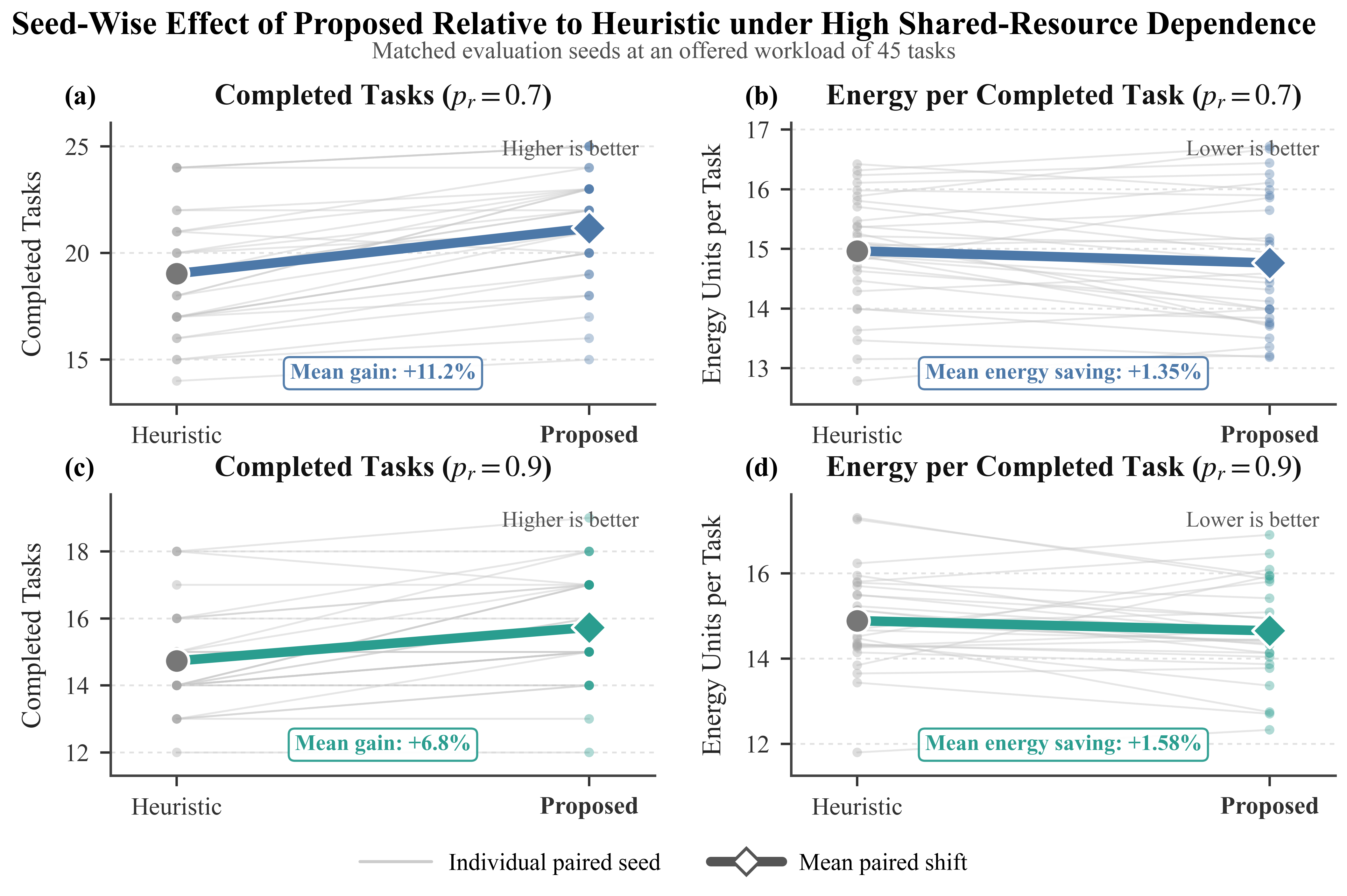}
\caption{Matched seed-wise changes from Heuristic to Proposed given 45 tasks: (a) completed tasks at $p_r=0.7$, (b) energy per completed task at $p_r=0.7$, (c) completed tasks at $p_r=0.9$, and (d) energy per completed task at $p_r=0.9$. Each thin gray line connects results obtained from the same held-out seed, while the highlighted line connects the corresponding method means. An upward change is favorable for completed tasks, whereas a downward change is favorable for energy per completed task.}
\label{fig:seedwise_results}
\end{figure}

Aggregate means do not show whether the observed differences are distributed broadly across stochastic realizations or driven by a small number of seeds. Fig.~\ref{fig:distributional_results} therefore reports the complete held-out distributions at $p_r=0.7$, where the three methods are compared using the same 30 simulation seeds. As shown in Fig.~\ref{fig:distributional_results}(a), the distribution of completed tasks for the proposed method is shifted upward relative to the heuristic and regression models. The tardiness distribution in Fig.~\ref{fig:distributional_results}(b) is shifted toward lower values. The deadline-miss distribution in Fig.~\ref{fig:distributional_results}(c) exhibits a lower central tendency for the proposed method, although its spread indicates that the magnitude of the benefit varies among stochastic realizations. In Fig.~\ref{fig:distributional_results}(d), the three energy distributions overlap, while our proposed method reduces the processing energy per completed task. To examine matched changes directly, Fig.~\ref{fig:seedwise_results} connects the heuristic and proposed outcomes obtained from the same evaluation seed. This paired representation removes between-seed differences in workload realizations and reveals the direction and magnitude of the change introduced by the ranking correction. Fig.~\ref{fig:seedwise_results}(a) and Fig.~\ref{fig:seedwise_results}(c) show that the higher mean number of completed tasks is supported by paired changes across the held-out seeds at both values of $p_r$. The energy panels show a downward shift in the mean. Completed-task gains are significant and the energy-per-task differences are favorable in direction.

\subsection{Sensitivity to Offered Workload}
\begin{figure}[t]
\centering
\includegraphics[width=0.98\columnwidth]{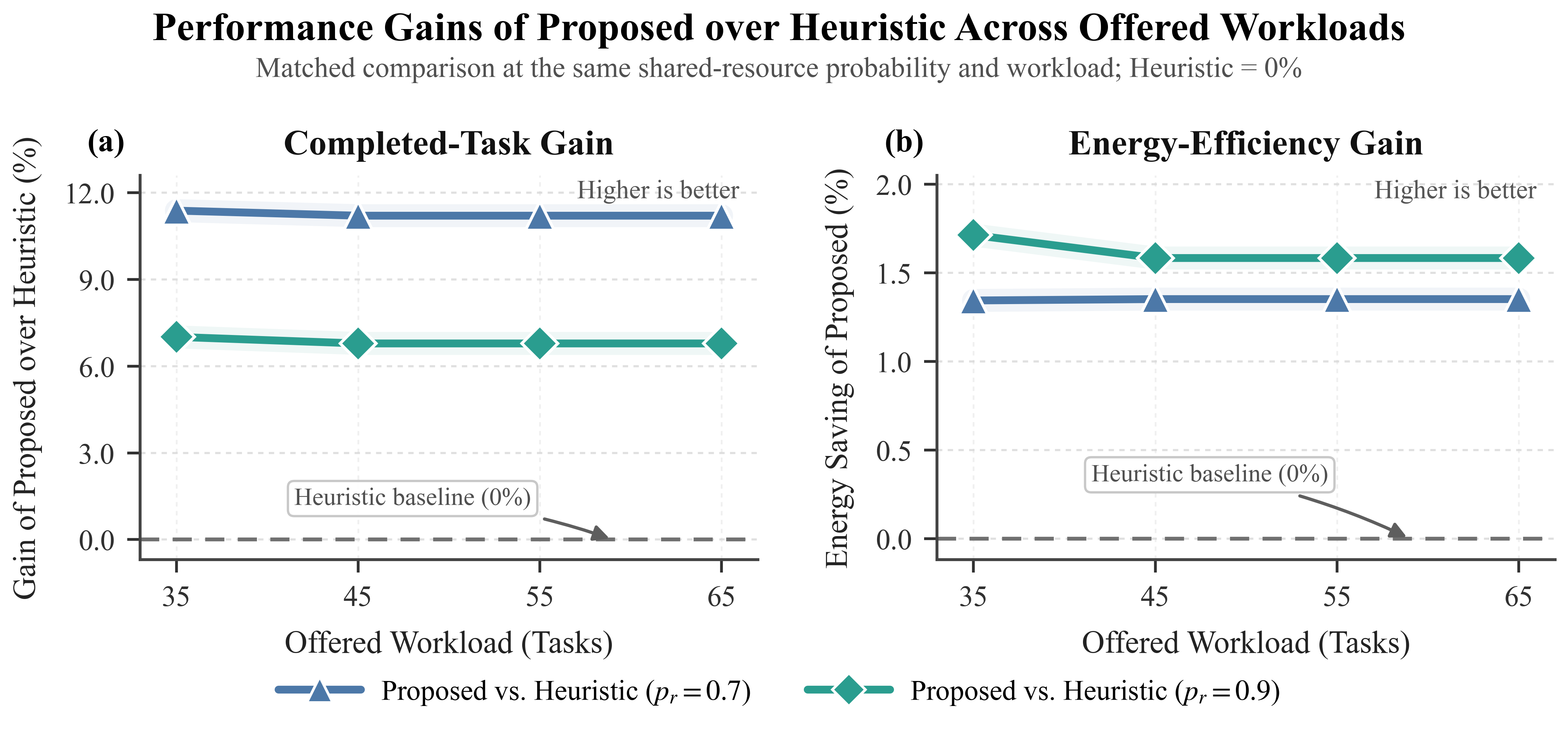}
\caption{Relative gains of Proposed over Heuristic across offered workloads of 35, 45, 55, and 65 tasks: (a) completed-task gain and (b) energy-efficiency gain. Positive values in both panels favor our proposed method, and the dashed zero line represents the Heuristic model.}
\label{fig:workload_sensitivity}
\end{figure}

The preceding results use a workload of 45 tasks. To determine whether the observed gains depend on this particular operating point, the offered workload is varied while the shared-resource requirement probability is fixed at either $p_r=0.7$ or $p_r=0.9$. For each workload, Proposed and Heuristic are evaluated under matched configurations and random seeds. Fig.~\ref{fig:workload_sensitivity}(a) shows that the proposed method maintains a positive completed-task gain throughout the evaluated workload range. The gain remains approximately stable within each shared-resource condition, with a larger improvement at $p_r=0.7$ than at $p_r=0.9$. This behavior indicates that the throughput advantage is not restricted to the primary workload of 45 tasks. Fig.~\ref{fig:workload_sensitivity}(b) similarly shows a consistently positive mean energy-efficiency gain across the evaluated workloads. This panel reports aggregate relative means and it should be interpreted together with the paired tests in Table~\ref{tab:primary_results}. Overall, the results show that the principal benefit of the proposed ranking correction lies in improving completed-task throughput and temporal performance when contention over a shared production asset is high. The regression baseline largely preserves the behavior of the analytical heuristic, whereas the pairwise ranking formulation changes the relative preference among candidate machines and therefore affects the winner-selection outcome. Under the evaluated circular manufacturing conditions, this produces more effective coordination of tasks that depend on a reusable, capacity-constrained asset. The observed energy-per-task behavior additionally suggests a favorable operational-efficiency effect.

\section{Conclusion}
\label{sec:conclusion}
This paper presents an Edge-AI-driven decentralized task-allocation framework for smart manufacturing systems operating under dynamic workloads and shared-resource constraints. The proposed mechanism retains an interpretable resource-aware analytical bid and augments it with a compact autoencoder-regularized pairwise ranking correction trained using counterfactual preference pairs. This design aligns the learned component with the ordering-based nature of minimum-bid winner selection while preserving lightweight machine-level inference and avoiding centralized global scheduling.

The experimental results demonstrate that our proposed mechanism improves completed-task throughput, average tardiness, and deadline adherence under high shared-resource dependence. These improvements are observed over fully held-out, seed-matched simulations and remain consistent across the evaluated offered workloads. The framework follows three progressive design stages. First, the resource-aware analytical heuristic establishes the decentralized negotiation structure. Second, an Edge-AI regression formulation examines whether learned approximation of the analytical bid improves allocation behavior. Finally, the proposed pairwise ranking correction explicitly targets candidate ordering, which directly governs decentralized winner selection. The regression formulation serves as an intermediate comparative stage and is not part of the final inference pipeline. The results show that regression largely preserves the allocation behavior of the analytical heuristic, whereas ranking produces more favorable task-machine ordering and system-level outcomes.

The proposed method also yields lower mean energy consumption per completed task. More broadly, the results show that ranking-aware negotiation can improve operational coordination when a large fraction of tasks competes for the same reusable, capacity-constrained production asset. This capability supports the operational resource-efficiency objectives of circular smart manufacturing. Since inference latency depends on the edge platform, runtime, and communication configuration, systematic cross-platform latency benchmarking remains necessary. Extending the deployment to more constrained embedded endpoints (i.e., microcontroller-class platforms) would require hardware-aware model optimization and implementation, such as reduced-precision inference, memory and buffer-aware execution, and device-level profiling of latency, energy consumption, and computational load under the target runtime and communication stack. Future work will extend the framework to multiple heterogeneous shared resources, larger machine configurations, direct latency benchmarking across edge processors and runtimes, and explicit material-circularity and resource-life-cycle indicators.



\ifCLASSOPTIONcaptionsoff
  \newpage
\fi

\vspace{-1.5\baselineskip}
\begin{IEEEbiography}[{\includegraphics[width=1in,height=1.25in,clip,keepaspectratio]{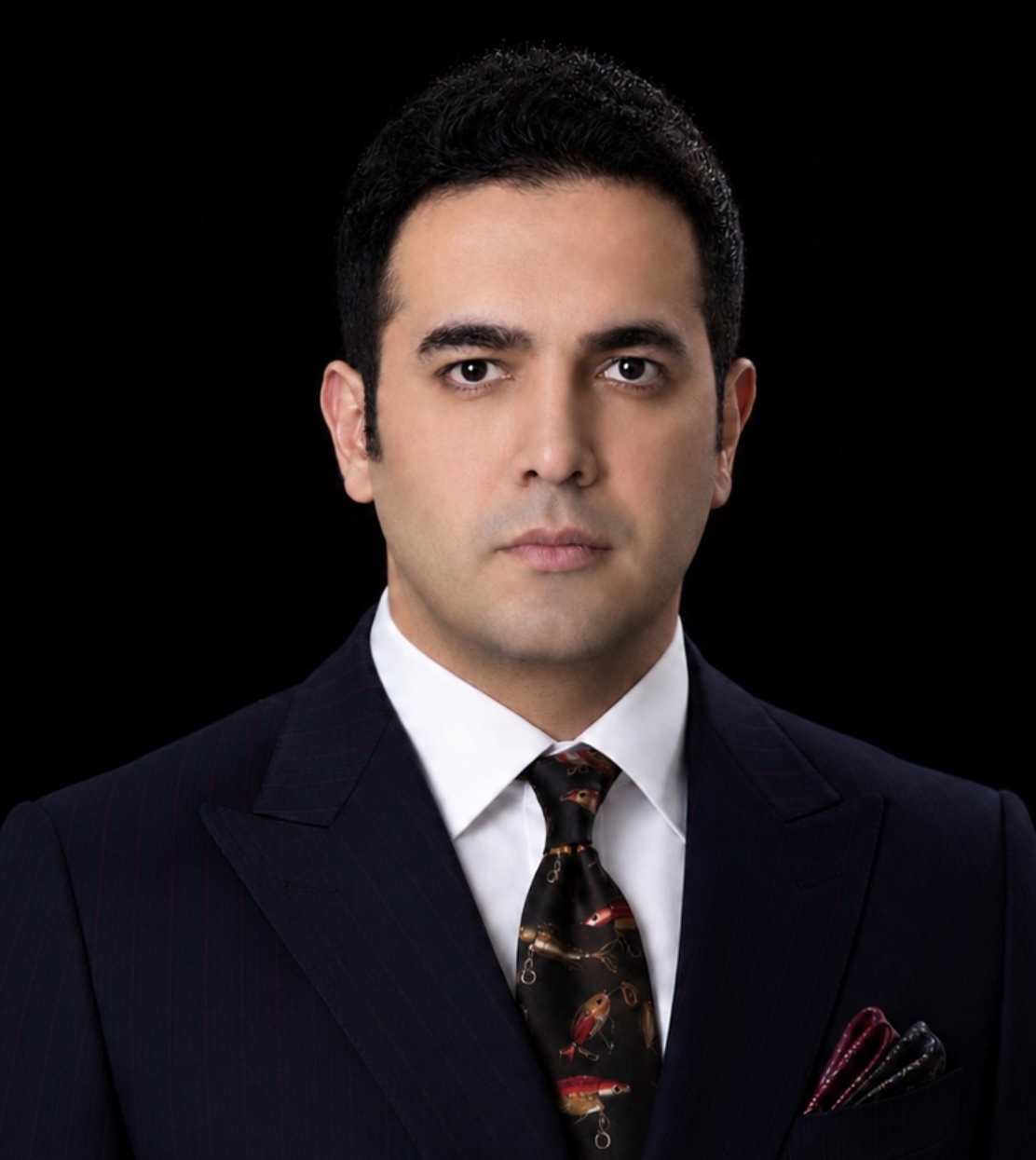}}]{Mohammadhossein Ghahramani}(Senior Member, IEEE) obtained his Ph.D. in Computer Technology and Application from Macau University of Science and Technology in 2018. He was a member of the Insight Centre for Data Analytics and a Research Fellow at University College Dublin, Ireland. Currently, he is an Assistant Professor of Data Science at Birmingham City University, UK. His research interests include smart systems, artificial intelligence, optimization, and IoT. Dr Ghahramani has published numerous papers in reputable journals and has received several awards. He serves as a co-chair of the IEEE SMCS Technical Committee on AI-based Smart Manufacturing Systems and as an Associate Editor of IEEE Internet of Things Journal.\end{IEEEbiography}

\vspace{-10pt}
\begin{IEEEbiography}[{\includegraphics[width=1in,height=1.25in,clip,keepaspectratio]{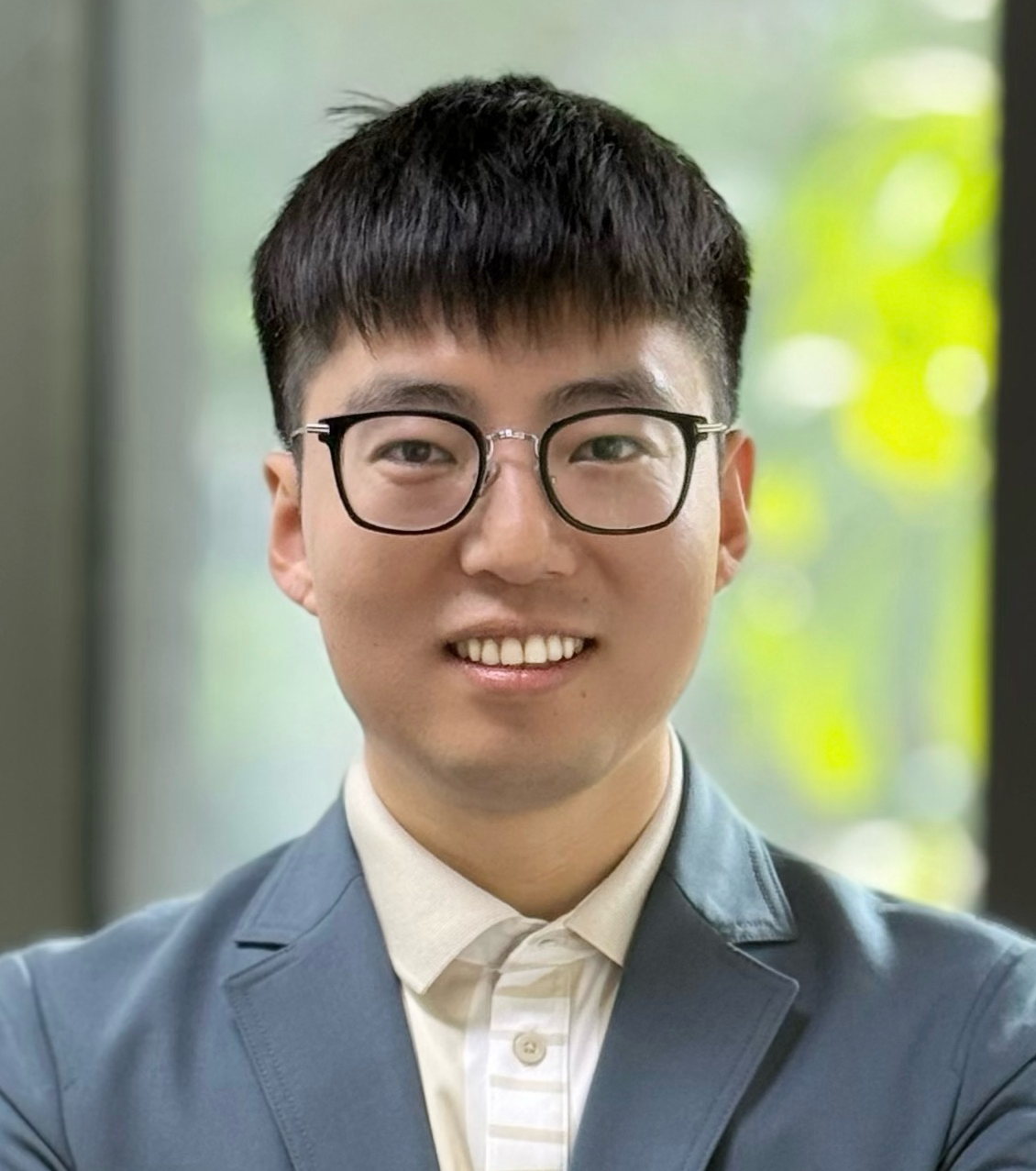}}]{Yan Qiao}(Senior Member, IEEE) received the B.S. and Ph.D. degrees in Industrial Engineering and Mechanical Engineering from Guangdong University of Technology, Guangzhou, China, in 2009 and 2015, respectively. He is currently a Professor at the College of Computer and Information Science, Southwest University, Chongqing, China. He is also the co-founder and Chief Scientist of IKAS Holdings (Beijing) Co., Ltd., China. His research interests include semiconductor smart manufacturing systems, scheduling and control, AI technologies and their applications. He has over 140 publications, including more than 60 papers in IEEE Transactions. He also has 17 patents (eight US patents). He has received several awards. Prof. Qiao serves as an Associate Editor of the IEEE Robotics and Automation Magazine.
\end{IEEEbiography}

\vspace{-10pt}
\begin{IEEEbiography}[{\includegraphics[width=1in,height=1.25in,clip,keepaspectratio]{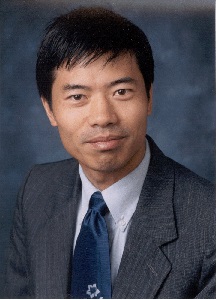}}]{MengChu Zhou}
(S'88-M'90-SM'93-F'03)(Fellow, IEEE) received his Ph. D. degree from Rensselaer Polytechnic Institute, Troy, NY in 1990 and then joined New Jersey Institute of Technology where he has been Distinguished Professor since 2013. His interests are in Petri nets, automation, robotics, big data, Internet of Things, cloud/edge computing, and AI. He has 1400+ publications including 18 books, 900+ journal papers (700+ in IEEE transactions), 32 patents and 32 book-chapters. He is Fellow of IFAC, AAAS, CAA and NAI.

\end{IEEEbiography}

\vfill
\vfill
\vfill


\end{document}